\documentclass{article}

\usepackage{microtype}
\usepackage{graphicx}
\usepackage{subfigure}
\usepackage{booktabs} 

\usepackage{hyperref}

\usepackage[archival]{icml2023}

\usepackage{amsmath}
\usepackage{amssymb}
\usepackage{mathtools}
\usepackage{amsthm}

\definecolor{airforceblue}{rgb}{0.36, 0.54, 0.66}
\definecolor{ao(english)}{rgb}{0.0, 0.5, 0.0}
\definecolor{arylideyellow}{rgb}{0.91, 0.84, 0.42}

\usepackage[capitalize,noabbrev]{cleveref}

\theoremstyle{plain}

\theoremstyle{definition}

\theoremstyle{remark}

\usepackage[textsize=tiny]{todonotes}

\icmltitlerunning{On Explicit Curvature Regularization in Deep Generative Models}

\begin{document}

\twocolumn[
\icmltitle{On Explicit Curvature Regularization in Deep Generative Models}




\begin{icmlauthorlist}
\icmlauthor{Yonghyeon Lee}{yyy}
\icmlauthor{Frank C. Park}{comp,lolo}
\end{icmlauthorlist}

\icmlaffiliation{yyy}{Korea Institute for Advanced Study, Seoul, South Korea}
\icmlaffiliation{comp}{Department of Mechanical Engineering, Seoul National University, Seoul, South Korea}
\icmlaffiliation{lolo}{Saige Research, Seoul, South Korea}

\icmlcorrespondingauthor{Yonghyeon Lee}{ylee@kias.re.kr}
\icmlcorrespondingauthor{Frank C. Park}{fcp@snu.ac.kr}

\icmlkeywords{Machine Learning, ICML}

\vskip 0.3in
]



\printAffiliationsAndNotice{}  


\begin{abstract}
We propose a family of curvature-based regularization terms for deep generative model learning.  Explicit coordinate-invariant formulas for both intrinsic and extrinsic curvature measures are derived for the case of arbitrary data manifolds embedded in higher-dimensional Euclidean space. Because computing the curvature is a highly computation-intensive process involving the evaluation of second-order derivatives, efficient formulas are derived for approximately evaluating intrinsic and extrinsic curvatures. Comparative studies are conducted that compare the relative efficacy of intrinsic versus extrinsic curvature-based regularization measures, as well as performance comparisons against existing autoencoder training methods. Experiments involving noisy motion capture data confirm that curvature-based methods outperform existing autoencoder regularization methods, with intrinsic
curvature measures slightly more effective than extrinsic curvature measures.
\end{abstract}


\section{Introduction}

For a large class of deep generative models, the problem of manifold learning
is central to their training: given a set of data points drawn from
some high-dimensional space ${\cal X}$ and for which the Manifold Hypothesis
remains valid -- that is, the data points are assumed to lie on some
lower-dimensional manifold ${\cal M} \subset {\cal X}$ -- the objective
is to determine a mapping $f: {\cal Z} \rightarrow {\cal X}$, where
${\cal Z}$ is typically a bounded region of a vector space with
the same dimension as ${\cal M}$, such that $f({\cal Z})$ closely
approximates ${\cal M}$. The mapping $f$ is the {\it generator} (or
the {\it decoder} in the context of autoencoders), ${\cal Z}$ is called
the {\it latent space}, and ${\cal M}$ is referred to as the {\it data
manifold}.

\begin{figure}[!t]
    \centering
    \includegraphics[width=0.5\textwidth]{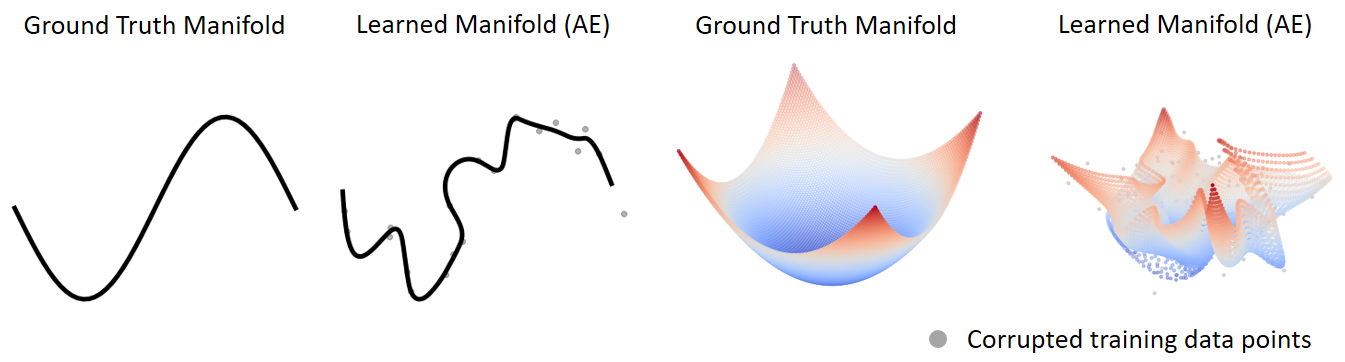}
    \vspace{-10pt}
    \caption{Autoencoder-based manifold learning examples given noise-corrupted training data.}
    \label{fig:intro_2}
    \vspace{-10pt}
\end{figure}

Deep neural networks are a popular choice of parametric mapping
for the generator $f$~\cite{lecun2015deep}. Methods for training
such deep generative models include those based on the
autoencoder~\cite{kramer1991nonlinear,kingma2013auto},
adversarial training~\cite{goodfellow2014generative}, 
manifold flow~\cite{brehmer2020flows}, and the autodecoder~\cite{bojanowski2017optimizing}.
As with any function approximation problem, the presence of noise in
the data can lead to overfitting, resulting in poor manifold approximations
of the type shown in Figure~\ref{fig:intro_2}.  Adding a regularization
term to the loss function is one means of mitigating the effects of
overfitting. With few exceptions the regularization terms are formulated
in terms of the gradient of $f$; the underlying intuition is
that by minimizing, e.g., the norm of the gradient, $f$ is forced to
be close to linear, and thus less prone to overfitting to any noise-induced 
variations in the data. 

Several recent works have pointed out the importance of formulating
both the loss function and regularization terms in a
{\it coordinate-invariant} way~\cite{2019riemannian, jang2020riemannian, lee2022regularized, jang2022geometrically, 2023geometric}.  
In fact, in \cite{lee2022regularized} a compelling case is made that the 
problem of determining $f$ is best
framed as one of finding a {\it minimum distortion} map between two
Riemannian manifolds: the key idea is to ``wrap'' ${\cal M}$ by ${\cal Z}$
in such a way that the overall distortion -- one can view ${\cal M}$ 
as being made of marble and ${\cal Z}$ of elastic, with the distortion
corresponding to the elastic energy resulting from deforming ${\cal Z}$ --
is minimized. This geometric framework allows for, among other things,
classifying existing manifold and representation learning approaches
based on, e.g., the choice of Riemannian metric, boundary conditions,
and other a priori specified factors.

In this paper we adopt a different geometric approach that focuses on
the {\it curvature} of the data manifold.  Intuitively, curvature
measures how much a manifold deviates from a flat manifold, and for
the purpose of learning a generator $f$ that is robust to noise,
a regularization term that attempts to minimize the curvature of
the resulting data manifold would seem quite sensible.  In the case
of classical two-dimensional surfaces embedded in $\mathbb{R}^3$,
curvature is quantified as the rate of change of the surface
normal along certain directions; for this, second-order derivatives
of the surface parametrization $f$ are involved. This is a
key fundamental difference with distortion-based regularization
terms, which involve only first-order derivatives of $f$.

\begin{figure}[!t]
    \centering
    \includegraphics[width=0.5\textwidth]{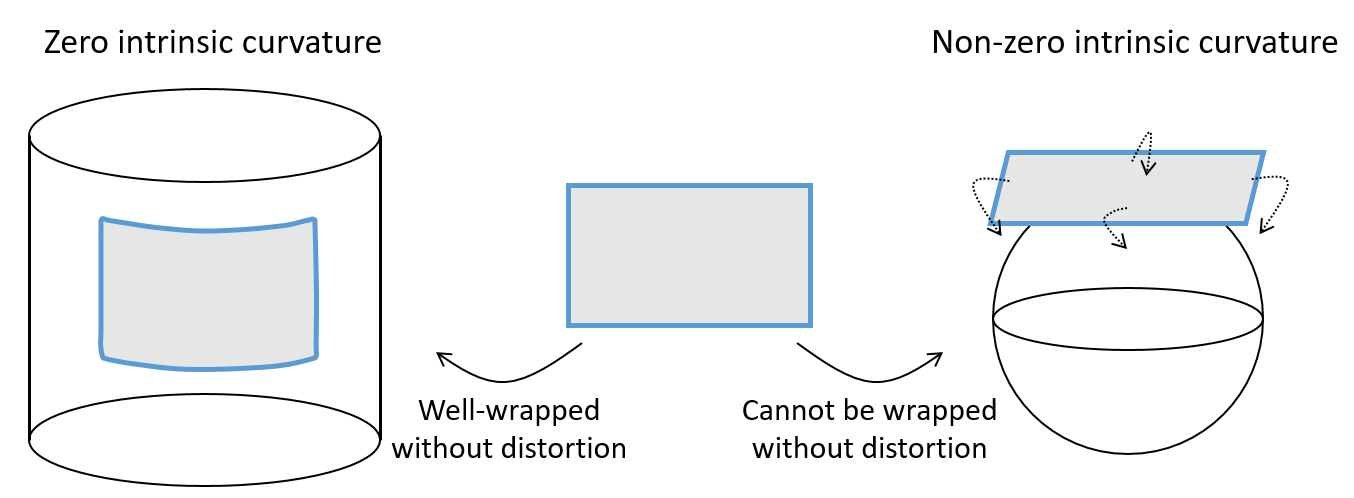}
    \vspace{-10pt}
    \caption{\textit{Left}: The cylindrical surface has zero intrinsic curvature but has non-zero extrinsic curvature. \textit{Right}: The spherical surface has non-zero intrinsic and extrinsic curvature.}
    \label{fig:vis_intrinsic_curvature}
    \vspace{-10pt}
\end{figure}

The theory of curvature for Riemannian manifolds is both well-developed
and at the same time intimidatingly complex, but the intuitive
ideas can be understood from the case of classical two-dimensional
surfaces~\cite{do1992riemannian, do2016differential}. The {\it
extrinsic curvature} of a surface captures how the surface is
embedded in $\mathbb{R}^3$, whereas the {\it intrinsic curvature}
is a property intrinsic to the surface, independent of its
embedding in $\mathbb{R}^3$. For example, a cylinder and a
flat piece of paper have different extrinsic curvatures
but identical intrinsic curvatures -- the flat paper can be
rolled into a cylinder without any deformations -- while a
sphere and cylinder have different intrinsic curvatures 
(see Figure~\ref{fig:vis_intrinsic_curvature}).

In the context of deep generative model learning, the role of
extrinsic versus intrinsic curvature in the formulation of a
regularization term has yet to be investigated. At first glance,
it would seem obvious that extrinsic measures are more 
relevant, since the shape of the data manifold -- specifically, how
it is embedded in the higher-dimensional ambient space -- is 
clearly important. On the other hand, generalizations of curvature to
higher-dimensional manifolds are focused almost exclusively
on intrinsic measures like the Ricci
curvature~\cite{ricci1904direzioni,besse2007einstein,
do1992riemannian}; coordinate-invariant measures of 
extrinsic curvature for higher-dimensional manifolds embedded
in Euclidean space have yet to be addressed in the 
literature.
Moreover, formulas for both intrinsic and extrinsic curvatures are 
heavily computation-intensive, as they involve second-order 
derivatives of the generator $f$.

This paper reports on three contributions.  First, we formulate
explicit coordinate-invariant extrinsic curvature measures for
multi-dimensional manifolds embedded in higher-dimensional
Euclidean space. The core idea is to generalize the Gauss map
for two-dimensional surfaces in $\mathbb{R}^{3}$ to higher-dimensional
embedded manifolds using the machinery of Grassmann
manifolds~\cite{wong1967differential, bendokat2020grassmann}
and Dirichlet energy~\cite{eells1978report, eells1988another}. 
Second, we derive computationally efficient algorithms
for approximately computing both intrinsic and extrinsic
curvature measures that are based on Hutchinson’s
stochastic trace estimator~\cite{hutchinson1989stochastic}.

The third and final contribution is a comparative study of
the efficacy of extrinsic versus intrinsic curvature measures
for generative deep model learning.  Adopting an autoencoder
framework, regularization terms that minimize intrinsic and 
extrinsic curvature measures are developed,
and applied to motion capture data.
Performance comparisons with existing autoencoder training
methods~\cite{vincent2010stacked, kingma2013auto, rifai2011contractive,
alain2014regularized, lee2021neighborhood, lee2022regularized} are also provided. 

A key finding of our study is that both intrinsic and extrinsic curvature-based regularization terms are more effective than existing autoencoder regularization methods in mitigating the effects of noise. The same can be said when comparing our curvature-based methods to purely first-order
gradient-based distortion measures. Intrinsic curvature measures, although requiring more computation than their extrinsic counterparts, appear to be slightly more effective. 


\section{Curvature of Riemannian Manifolds}

In what follows we consider an $m$-dimensional Riemannian manifold
$\mathcal{M}$ embedded in $\mathbb{R}^{D}$. Choosing $z \in \mathbb{R}^m$
as local coordinates for ${\cal M}$, ${\cal M}$ is parametrized
by the mapping $f: {\mathbb{R}^m} \rightarrow {\cal M}$, i.e., $z
\mapsto x=f(z)$. Here $x \in {\cal M} \subset \mathbb{R}^D$.
The Riemannian metric on ${\cal M}$ is obtained in terms of local
coordinates $z$ as $G(z):=J_f^T(z) J_f(z)$ where $J_f(z)=
\frac{\partial f}{\partial z}(z)$ (this follows from the Euclidean
incremental arclength formula $ds^2 = dx_1^2 + \ldots dx_D^2 =
dx^T dx = dz^T J_f^T J_f dz$, where
$dx = \frac{\partial f}{\partial z} dz = J_f dz$). 
The $(i,j)$-th component of the Riemannian metric $G$ is denoted by 
$g_{ij}$, and that of $G^{-1}$ is denoted by $g^{ij}$.

\subsection{Intrinsic Curvature}
For classical two-dimensional surfaces in $\mathbb{R}^3$, the Gaussian curvature, which can be obtained as the product of the principal curvatures, is an intrinsic curvature measure. For higher-dimensional Riemannian manifolds, the study of intrinsic curvature begins with the Riemann curvature tensor~\cite{do1992riemannian}, which is defined in coordinates as
\begin{equation}
    R^{a}_{bcd} = \frac{\partial}{\partial z^{c}}\Gamma^{a}_{db} - \frac{\partial}{\partial z^{d}}\Gamma^{a}_{cb} + 
    \sum_{\lambda=1}^{m} \Gamma^{a}_{c\lambda}\Gamma^{\lambda}_{db} - \sum_{\lambda=1}^{m} \Gamma^{a}_{d\lambda}\Gamma^{\lambda}_{cb},
\end{equation}
where $\Gamma^{a}_{bc}$ are the Christoffel symbols of the second kind:
\begin{equation}
\label{eq:christoffel}
    \Gamma^{a}_{bc} = \frac{1}{2} \sum_{\lambda=1}^{m} g^{a\lambda}(\frac{\partial}{\partial z ^c}g_{\lambda b} + \frac{\partial}{\partial z^b}g_{\lambda c} - \frac{\partial}{\partial z^{\lambda}}g_{b c}).
\end{equation}
The Ricci curvature is obtained by contracting the Riemann curvature tensor:  $\mathrm{Ric}_{ij}(z):=\sum_{a=1}^{m}R^{a}_{iaj}(z)$. In the case of two-dimensional surfaces, the Ricci curvature reduces to $\mathrm{Ric}_{ij}(z) = K g_{ij}(z)$ where $K$ is the Gaussian curvature.

The scalar curvature $R=\sum_{i,j} \mathrm{Ric}_{ij}(z)g^{ij}(z)$ is a natural scalar measure of the intrinsic curvature of a Riemannian manifold. We note that $R=2K$ for two-dimensional surfaces. Just like Gaussian curvature, the scalar curvature can be negative or positive. In what follows we use the squared scalar curvature as a local intrinsic curvature measure:
\begin{equation}
\label{eq:ICmeasure}
    \textrm{Intrinsic Curvature}(z)=\big( \mathrm{Tr}(G^{-1}(z)\mathrm{Ric}(z)) \big)^2.
\end{equation}
More detailed and formal treatments of the Riemann curvature tensor can be found in, e.g.,~\cite{do1992riemannian}. A more intuitive derivation of the curvature tensor based on the notion of parallel transport can be found in~\cite{fecko2006differential, schutz2022first}.

\subsection{Extrinsic Curvature}
For a two-dimensional surface ${\cal S}$ in $\mathbb{R}^{3}$, its extrinsic curvature can be quantified as the rate of change of the surface normal. Specifically, the Gauss map $n:\mathcal{S} \to \mathbb{S}^{2}$ assigns to each point $x\in\mathcal{S}$ a unit vector $n(x)$ that is normal to $\mathcal{S}$ at $x$. The differential of the Gauss map then leads to a measure of the extrinsic curvature via the second fundamental form~\cite{do2016differential, kuhnel2015differential}. 

\begin{figure}[!t]
    \centering
    \includegraphics[width=0.45\textwidth]{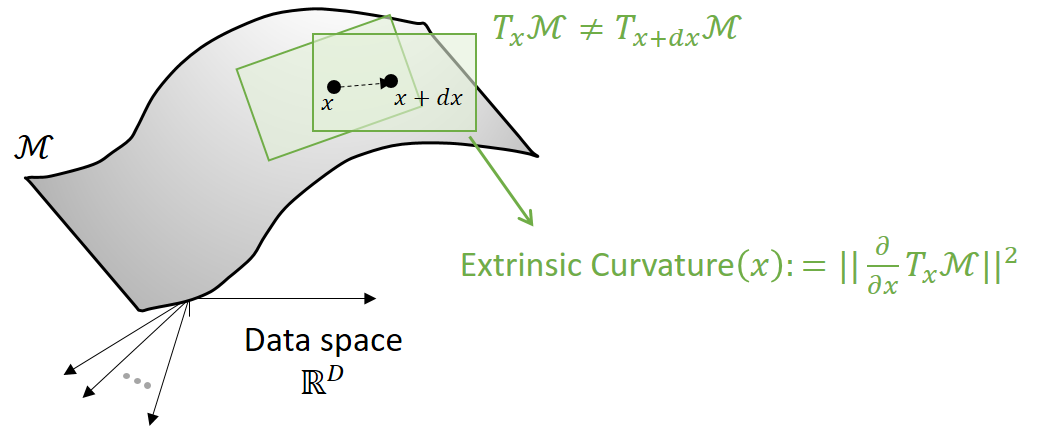}
    \caption{The tangent space to $\mathcal{M}$ at $x$ varies as $x$ moves to $x+dx$, i.e., $T_{x}\mathcal{M}\neq T_{x+dx}\mathcal{M}$; the manifold has non-zero extrinsic curvature at $x$.}
    \label{fig:vis_extrinsic_curvature}
    \vspace{-10pt}
\end{figure}

We now generalize the above Gauss map construction to an $m$-dimensional manifold $\mathcal{M}$ in $\mathbb{R}^{D}$ and formulate a coordinate-invariant extrinsic curvature measure as the norm of $\frac{\partial }{\partial x} T_x \mathcal{M}$, where $T_x \mathcal{M}$ denotes the tangent space to $\mathcal{M}$ at $x$ (see Figure~\ref{fig:vis_extrinsic_curvature}). The first step is to establish a 1-1 correspondence between the set of tangent spaces and the Grassmann manifold $\text{Gr}(m, \mathbb{R}^D)$:
\begin{equation}
\small
    \text{Gr}(m, \mathbb{R}^{D}) = \{P \in \mathbb{R}^{D\times D} \ | \ P^T=P, \ P^2=P, \ \text{rank}(P)=m\}.
\label{eq:grassmann}
\end{equation}
$\text{Gr}(m, \mathbb{R}^D)$ can be viewed as the $m(D-m)$-dimensional manifold of orthogonal projection matrices; every element $P \in \text{Gr}(m, 
\mathbb{R}^{D})$ can be uniquely identified with the linear subspace $\text{Range}(P)\subset \mathbb{R}^{D}$~\cite{bendokat2020grassmann}\footnote{This is an implicit parametrization of the Grassmann manifold viewed as being embedded in Euclidean space $\mathbb{R}^{D\times D}$.}.
For a generator $f:\mathbb{R}^{m} \to \mathcal{M}$, the tangent space of $\mathcal{M}$ at $x=f(z)$ is equal to the range of $J_f(z)$, and hence the orthogonal projection matrix $E(J_f):=J_f(J_f^T J_f)^{-1} J_f^T$, which is an element of $\text{Gr}(m, \mathbb{R}^{D})$, can be identified with the tangent space $T_{x}\mathcal{M}$ in a 1-to-1 manner. We note that the representation $E(J_f)$ is coordinate-invariant; the proof is given in Appendix~\ref{appendix:cioecm}. 

We next consider a generalization of the Gauss map. The mapping $\text{T}:\mathcal{M} \to \text{Gr}(m,\mathbb{R}^{D})$ assigns to each point $x\in\mathcal{M}$ an $m$-dimensional linear subspace in $\mathbb{R}^{D}$ tangential to $\mathcal{M}$. In local coordinates $z$, this mapping can be expressed as
\begin{equation}
    T(z) = E(J_f(z)) = J_f(z)(J_f(z)^T J_f(z))^{-1} J_f(z)^T.
\end{equation}
Finally, taking the standard Riemannian metric on the Grassmann manifold given by
\begin{equation}
\langle V_1, V_2 \rangle := \text{Tr}(V_1^T V_2)
\label{eq:grassman_metric}
\end{equation}
for any $ V_1,V_2 \in T_{P}\text{Gr}(m,\mathbb{R}^D)$, 
we use the Dirichlet energy of  the mapping $\text{T}$~\cite{eells1978report, eells1988another}, which is a natural coordinate-invariant smoothness measure for a mapping between two Riemannian manifolds, to define the local extrinsic curvature measure as follows:
\begin{equation}
\label{eq:local_extrinsic_curvature}
\small
    \text{Extrinsic Curvature}(z):=\frac{1}{2}\text{Tr}(\sum_{i,j=1}^{m} (G^{-1})_{ij} \big( \frac{\partial }{\partial z^i} T \big)^T \frac{\partial }{\partial z^j} T).
\end{equation}
This measure is coordinate-invariant; the proof is given in Appendix~\ref{appendix:cioecm}.

\section{Minimum Curvature Deep Generative Models}

In this section we propose a curvature minimization framework for deep
generative models. We first develop efficient methods for estimating
the local intrinsic and extrinsic curvature measures in
Section~\ref{subsec:ice} and Section~\ref{subsec:ece}.
Finally, we explain how our local curvature measures can be used in deep
generative models in Section~\ref{subsec:mcml}.    

Throughout we will use Hutchinson's stochastic trace
estimator~\cite{hutchinson1989stochastic} to simplify the
above curvature measures: for an $n\times n$ matrix
$A$, $\text{Tr}(A) = \mathbb{E}_{v\sim \mathcal{N}(0, I_n)}[v^T A v]$,
where $I_n$ denotes the $n \times n$ identity matrix.  For the
purposes of this paper, unless otherwise specified
we set the number of samples from
$\mathcal{N}(0, I_n)$ in the trace estimation to be always $1$
in the training phase, i.e.,
$\text{Tr}(A) \approx v^T A v$ for $v\in \mathcal{N}(0, I_n)$.
We use the Einstein summation notation~\cite{einstein1922general}, i.e.,
when an index variable appears twice in a single term, it implies the
summation of that term over all the values of the index, e.g., $v^i_i
= \sum_{i}v^i_i$.  Finally, we use the shorthand notation 
$\partial_i := \frac{\partial }{\partial z^i}$.

\subsection{Intrinsic Curvature Approximation}
\label{subsec:ice}
Using Hutchinson’s trace estimator, we can estimate the local intrinsic curvature in~(\ref{eq:ICmeasure}) as $(v^T G^{-1}(z)\text{Ric}(z) v)^2$ for $v\in \mathcal{N}(0, I_m)$. 
Denoting $G^{-1}(z) v$ by $\tilde{v}$, the square root of the estimate is then
\begin{equation}
\label{eq:est_ic_measure}
    \tilde{v}^{T} \text{Ric}(z) v = (\partial_a \Gamma^{a}_{ij} - \partial_{j}\Gamma^{a}_{ai} + \Gamma^{a}_{ab}\Gamma^{b}_{ij} - \Gamma^{a}_{ib}\Gamma^{b}_{aj})\tilde{v}^{i}v^{j}. 
\end{equation}
Computing the Christoffel symbols $\Gamma^{i}_{jk}$ for all $i,j,k$, and using these to compute~(\ref{eq:est_ic_measure}) is practically infeasible due to memory limitations.  Instead, approximate formulas for the four terms in~(\ref{eq:est_ic_measure}) are derived that can be computed using the Jacobian-vector and vector-Jacobian products. 
We assume that $\partial \tilde{v}=\partial(G^{-1}v)$ is small enough to ignore and derive a more compact expression ($\partial G^{-1} = -G^{-1} \partial G G^{-1}$, and $\|G^{-1}\|$ is often very small as the norm of $J_f$ is typically large for high-dimensional data). 
Our later experiments show that the approximate intrinsic curvature with this assumption can effectively reduce the actual intrinsic curvature (Figure~\ref{fig:syn1} and \ref{fig:NTU_results}).
For space limitations, we derive the approximate formula only for the first term in~(\ref{eq:est_ic_measure}) only; formulas for the remaining terms are given in Appendix~\ref{appendix:licef}.

Substituting (\ref{eq:christoffel}) into the first term of (\ref{eq:est_ic_measure}), we get
\begin{equation}
\label{eq:ic_first_term_deta}
    \frac{1}{2}\partial_{a} \big( g^{a\lambda} ( v^j\partial_j (g_{\lambda i}\tilde{v}^i) + \tilde{v}^i\partial_i (g_{\lambda j}v^j) - \partial_{\lambda} (\tilde{v}^i g_{ij} v^j)) \big). 
\end{equation}
Since $\tilde{v}^ig_{ij} = v_i$, the first and third terms in (\ref{eq:ic_first_term_deta}) vanish. The first term in (\ref{eq:est_ic_measure}) then simplifies to
\begin{equation}
    \partial_a \Gamma^{a}_{ij}\tilde{v}^{i}v^{j} = \frac{1}{2} \text{Tr} \big( \nabla ( G^{-1} (\tilde{v}\cdot \nabla) (Gv)) \big).
\end{equation}
Again using Hutchinson’s trace estimator,
\begin{equation}
\small
    \frac{1}{2}\text{Tr} \big( \nabla ( G^{-1} (\tilde{v}\cdot \nabla) (Gv) ) \big) \approx \frac{1}{2} w^T \big( (w\cdot\nabla) ( G^{-1} (\tilde{v}\cdot \nabla) (Gv) ) \big) 
\end{equation}
for $w \in \mathcal{N}(0, I_m)$. The final approximate formula can be 
computed by repeatedly using the Jacobian-vector and vector-Jacobian products.
The remaining terms in (\ref{eq:est_ic_measure}) can be computed similarly;
see Appendix~\ref{appendix:licef}.   
The full expression for the estimated local intrinsic curvature is given as follows:
\begin{align}
\label{eq:eis_formula_all}
    &\Big( \frac{1}{2} (w \cdot \nabla) (w^T G^{-2} (v\cdot \nabla) (Gv)) \nonumber \\
    & - \frac{1}{2} (v \cdot \nabla) (w^T G^{-2} (v\cdot \nabla) (Gw)) \nonumber \\
    & + \frac{1}{4} (w^T G^{-3} (v\cdot \nabla)(G) (v\cdot\nabla)(Gw)) \nonumber \\
    & - \frac{1}{4} (w^T G^{-2} (v\cdot \nabla)(G) G^{-1} (v\cdot\nabla)(Gw)) \nonumber \\
    & - \frac{1}{4} (w^T G^{-2} (v\cdot \nabla)(G) G^{-1} (w\cdot\nabla)(Gv)) \nonumber \\ 
    & + \frac{1}{4}(w^T G^{-1} (v\cdot \nabla)(G) G^{-2} (w\cdot\nabla)(Gv)) \Big)^2,
\end{align}
where $v,w \in \mathcal{N}(0, I)$ and $G=J_f^T J_f$; it can be computed by using the Jacobian-vector and vector-Jacobian products multiple times.

\subsection{Extrinsic Curvature Approximation}
\label{subsec:ece} 
Using Hutchinson’s trace estimator, we can estimate the local extrinsic curvature in (\ref{eq:local_extrinsic_curvature}) as 
\begin{align}
    \frac{1}{2}v^Tg^{ij}\partial_i T \partial_j T v &=  \frac{1}{2}g^{ij}\partial_i (v_k T^k_l) \partial_j (T^l_m v^m) \nonumber \\ 
     &= \frac{1}{2}\text{Tr}( (\nabla (Tv))^T \nabla(Tv) G^{-1}).
\end{align}
for $v\in \mathcal{N}(0, I_D)$. Again via Hutchinson’s trace estimator, we get
\begin{equation}
\label{eq:esti_extrinsic_curv}
\small
   \frac{1}{2} w^T (\nabla (Tv))^T \nabla(Tv) G^{-1}w = \frac{1}{2}\big( (w \cdot \nabla) (Tv) \big)^T (\tilde{w} \cdot \nabla) (Tv), 
\end{equation}
for $w\in \mathcal{N}(0, I_m)$ and $\tilde{w} = G^{-1} w$; we can compute the final approximate formula via repeated use of the Jacobian-vector and vector-Jacobian products. 

\subsection{Curvature Minimization Framework}
\label{subsec:mcml}
In this section, we describe our curvature minimization framework for
deep generative model learning.  Essentially, we integrate the local
curvature measures from Section~\ref{subsec:ice}, \ref{subsec:ece} 
to define global curvature measures which are then
augmented to each of the original loss functions used in existing
deep manifold learning methods.  It is not necessary to integrate
these measures over the entire latent space $\mathbb{R}^{m}$. 
Rather, a probability density in the latent space $p(z)$ is
assumed available.  For GAN~\cite{goodfellow2014generative}
and manifold flow~\cite{brehmer2020flows}, the sampling distribution
is defined in the latent space.  For
autoencoders~\cite{kramer1991nonlinear,kingma2013auto}, given
an encoder $g:\mathbb{R}^{D}\to\mathbb{R}^{m}$, the pushforward
of the data distribution by $g$ is defined in the latent space
(i.e., as the aggregated posterior).  For the
autodecoder~\cite{bojanowski2017optimizing}, there exists a set
of latent vectors simultaneously optimized, which can then be
used to construct $p(z)$.

We define the global curvature measure as an expectation of
the local curvature measure over $p(z)$, which is then multiplied
by the weight parameter $\alpha$ and added to the original loss. 
Further algorithmic details can be found in Appendix~\ref{appendix:algo_details}.

\section{Experiments}
In this section, with applications of our minimum curvature framework to autoencoders, we implement Minimum Intrinsic Curvature Autoencoder (MICAE) and Minimum Extrinsic Curvature Autoencoder (MECAE). 
Our minimum curvature autoencoders are compared to the existing methods: vanilla Autoencoder (AE)~\cite{kramer1991nonlinear}, Variational Autoencoder (VAE)~\cite{kingma2013auto}, Contractive Autoencoder (CAE)~\cite{rifai2011higher}, De-noising Autoencoder (DAE)~\cite{vincent2010stacked}, Reconstruction Contractive Autoencoder (RCAE)~\cite{alain2014regularized}, Isometrically Regularized Autoencoder (IRAE)~\cite{lee2022regularized}, and Neighborhood Reconstructing Autoencoder (NRAE)~\cite{lee2021neighborhood}. 

Throughout, we assume that clean data is not available during training. 
Among models trained with various hyper-parameters, 
we select the one with the lowest mean validation reconstruction error. 
In this selection process we use clean validation data since the ability to reconstruct corrupted data well is not a desirable property of good models.

In section~\ref{subsec:synthetic data}, with a synthetic two-dimensional manifold example, we empirically show that, as the level of noise added to the training data increases, AE tends to learn more curved manifolds in both intrinsic and extrinsic senses, and minimizing curvatures with the proposed methods helps to learn more accurate manifolds.   

In section~\ref{subsec:skeleton}, we train ours and existing autoencoders with human skeleton pose data corrupted by Gaussian noise and compare the manifold learning performance. 
To evaluate quantitatively, we compare two mean test reconstruction errors. First, \textit{clean2clean} measures the mean reconstruction error of the clean test data. Secondly, \textit{corrupt2clean} measures the mean square error between the reconstruction of the corrupted test data and clean test data.
Experimental details not visible in the main script (e.g., network architectures) and comparisons of computational costs can be found in Appendix~\ref{appendix:exp_details}. 

\subsection{Synthetic Data}
\label{subsec:synthetic data}
\begin{figure}[!t]
    \centering
    \includegraphics[width=0.5\textwidth]{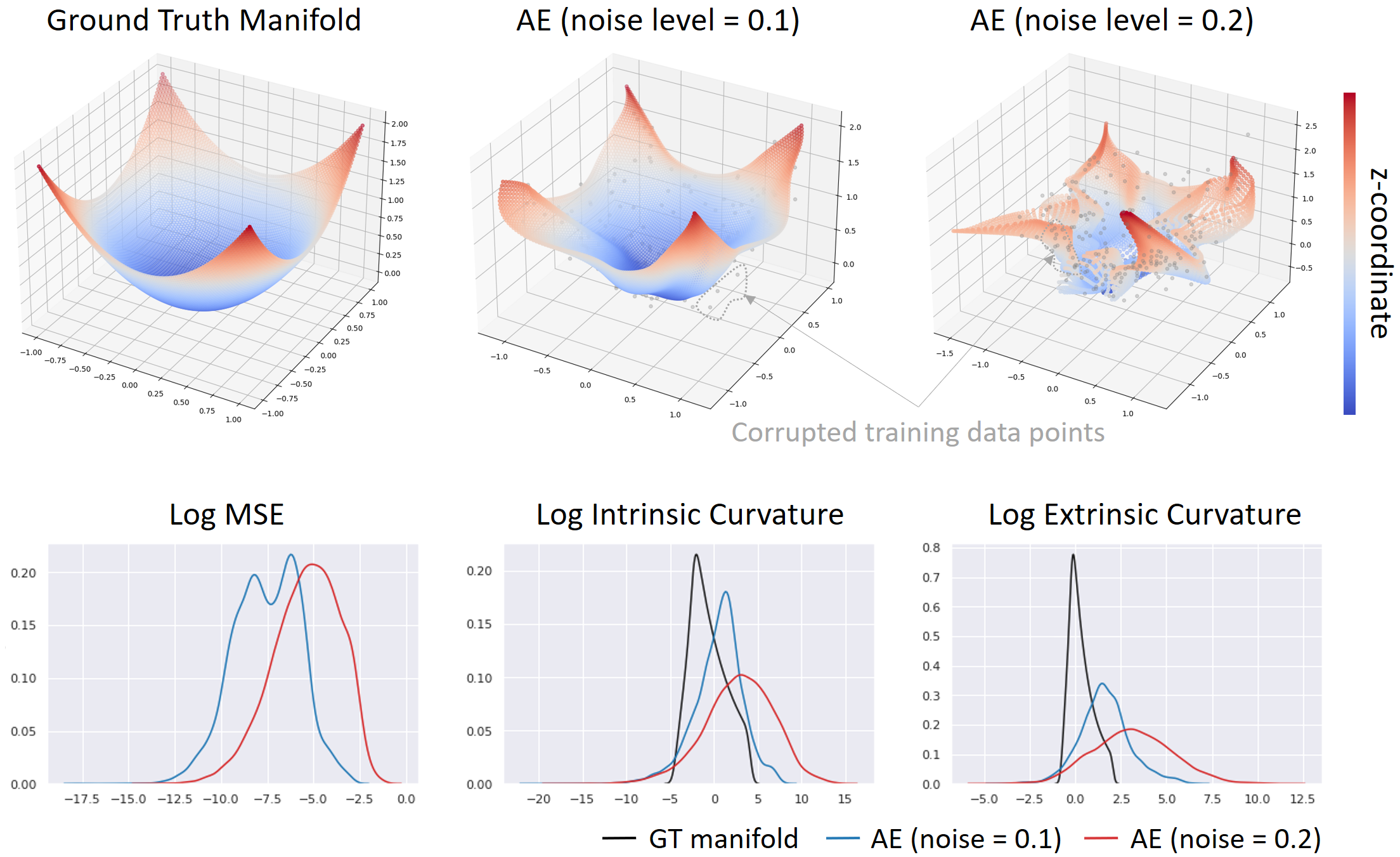}
    \vspace{-10pt}
    \caption{\textit{Upper}: Ground truth manifold, and learned manifolds by AE given noisy training data sets (the noise level means the standard deviation). \textit{Lower}: Density plots for the log MSE, Log Intrinsic Curvature, and Log Extrinsic Curvature. Log MSEs for the GT manifold are always $-\infty$.}
    \label{fig:syn1}
\end{figure}

Consider a ground truth two-dimensional manifold embedded in $\mathbb{R}^{3}$, that is $\mathcal{M}:=\{(x,y,x^2+y^2) \ | \ x,y \in (-1, 1)\}$; see Figure \ref{fig:syn1} (\textit{Upper Left}). 
We sample 200 random points in $(x,y)$-space, map them to $\mathbb{R}^{3}$, and add Gaussian noise to construct the training data set. 
We use three-layer fully connected neural networks (512 nodes per layer) for both encoder and decoder with ELU activation functions, and the latent space dimension is 2. 
The test data set is constructed with the $100 \times 100$ two-dimensional mesh grid in $(x,y)$-space (no noise added). 

Figure~\ref{fig:syn1} demonstrates how AE learns manifolds as the level of noise added to the training data increases. 
As the noise level increases, AE learns manifolds with higher intrinsic and extrinsic curvatures, as shown in the log curvature plots, that are less accurate, as shown in the log MSE density plot (the lower, the more accurate).

\begin{figure}[!t]
    \centering
    \includegraphics[width=0.4\textwidth]{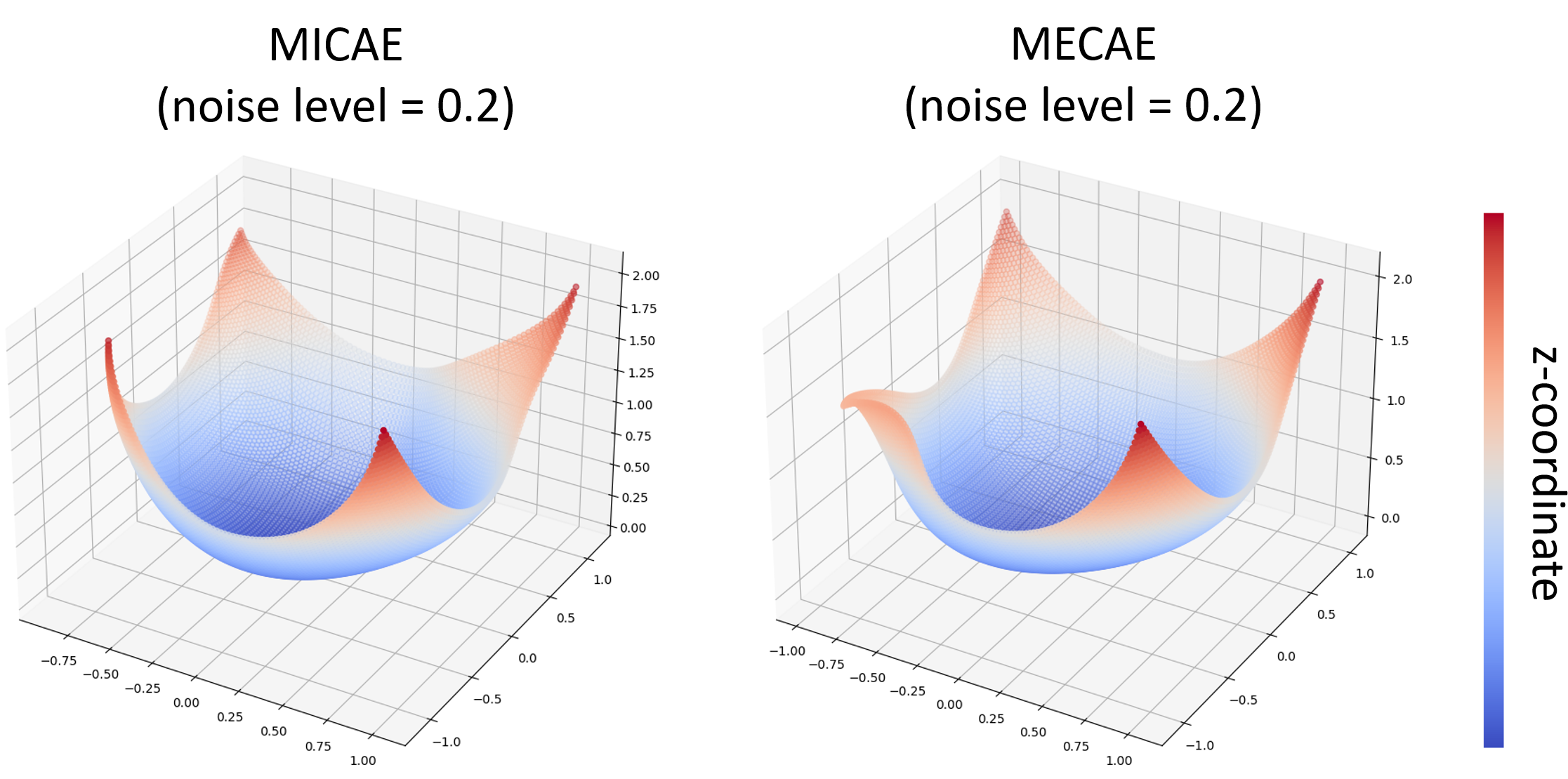}
    \vspace{-10pt}
    \caption{Learned manifolds by minimum curvature autoencoders given noisy training data sets.}
    \label{fig:syn2}
    \vspace{-5pt}
\end{figure}

Figure~\ref{fig:syn2} shows the manifold learning results of the proposed minimum curvature autoencoders trained with the same training data as in Figure~\ref{fig:syn1} (\textit{Upper Right}) (the noise level is $0.2$), and the density plots compared to the vanilla AE. 
Overall, our methods learn flatter and more accurate manifolds than AE.

\begin{figure}[!t]
    \centering
    \includegraphics[width=0.5\textwidth]{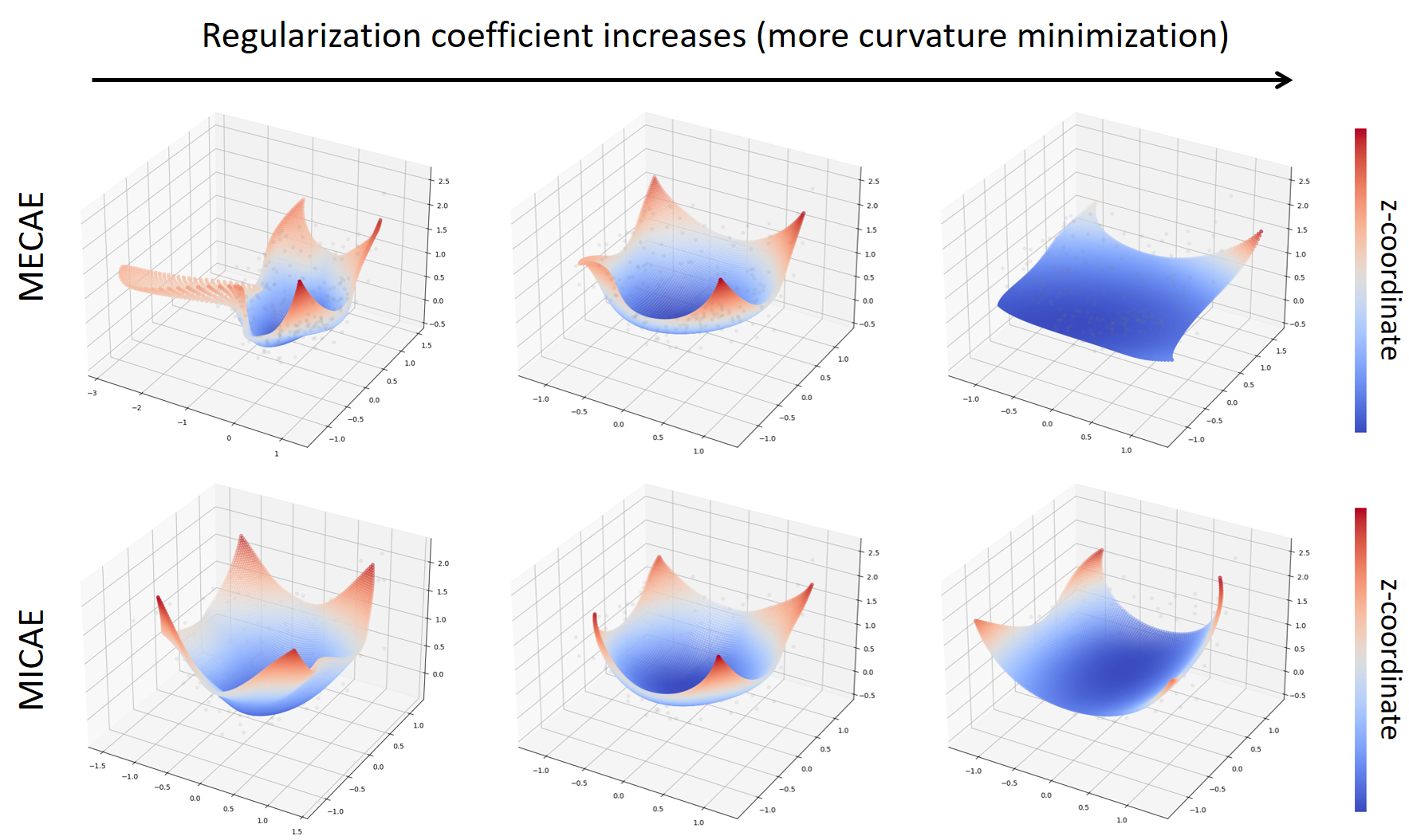}
    \vspace{-10pt}
    \caption{Illustration of how the learned manifold varies as the regularization coefficient increases. \textit{Upper}: MECAE. \textit{Lower}: MICAE. }
    \label{fig:syn4}
\end{figure}

Figure~\ref{fig:syn4} shows how the learned manifold varies as the regularization coefficient increases. 
In both MICAE and MECAE, if the regularization coefficient is too small, the manifold is still overfitting to noise. If it is too large, the curvature is excessively reduced, and the manifold becomes inaccurate. We should choose an appropriate level of regularization coefficient. 
As the regularization coefficient increases, manifolds learned by MECAE and MICAE converge differently.
While the MECAE learns a plane-like manifold that is extrinsically flat, the MICAE learns a cylinder-like manifold that is intrinsically flat. 

\begin{figure}[!t]
    \centering
    \includegraphics[width=0.45\textwidth]{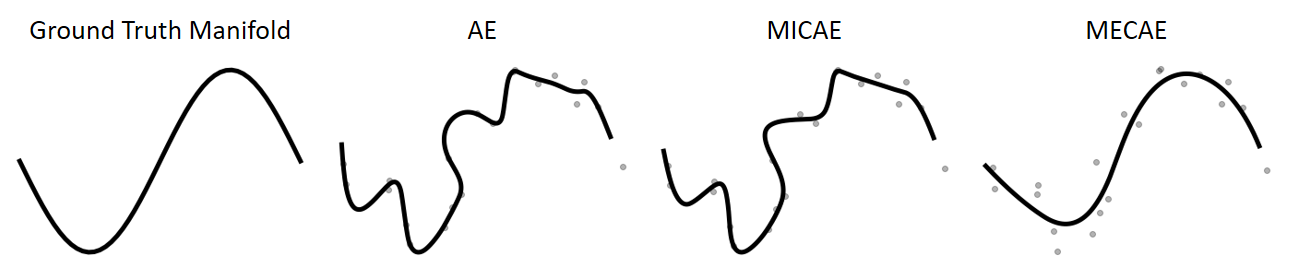}
    \vspace{-5pt}
    \caption{One-dimensional manifold learning results.}
    \label{fig:syn3}
\end{figure}

Figure~\ref{fig:syn3} shows one-dimensional manifold learning results trained with the noise-corrupted data, where the ground truth manifold is a sin-curve manifold. Since the intrinsic curvature of a curve is always zero, the intrinsic curvature minimization in MICAE does not affect manifold learning. On the other hand, MECAE learns a smoother and more accurate manifold.  

\textbf{Comparison to First-Order Distortion Minimization:} 
Isometrically Regularized Autoencoder (IRAE)~\cite{lee2022regularized}, which is a first-order distortion minimization framework, 
attempts to make the pullback Riemannian metric $G(z)=J_f^T(z)J_f(z)$ be proportional to the identity, i.e., $G(z)=cI$ for some $c>0$ and for all $z\in\mathcal{Z}$. In other words, the generator $f$ is regularized to be a scaled isometry that preserves angles and scaled distances between the Euclidean latent space and the learned data manifold. 
Interestingly, this distortion minimization approach has a close relation to intrinsic curvature minimization.

According to Gauss's Theorema Egregium (i.e., Gauss's Remarkable Theorem), the Gaussian curvature of a surface is invariant under local isometry. In other words, if two surfaces or manifolds are mapped to each other without distortion, then their Gaussian curvatures or intrinsic curvatures should be preserved. 
Therefore, if $f$ is a scaled isometry, then since the intrinsic curvature of the Euclidean latent space is everywhere 0, the resulting manifold's intrinsic curvature must be 0 everywhere as well.
As a result, in IRAE, intrinsic curvatures are implicitly minimized as a byproduct of distortion minimization, whereas MICAE directly minimizes intrinsic curvatures. 
In the following, we compare IRAE and MICAE with the above synthetic manifold example with a noise level of 0.2.

\begin{figure}[!h]
    \centering
    \includegraphics[width=0.4\textwidth]{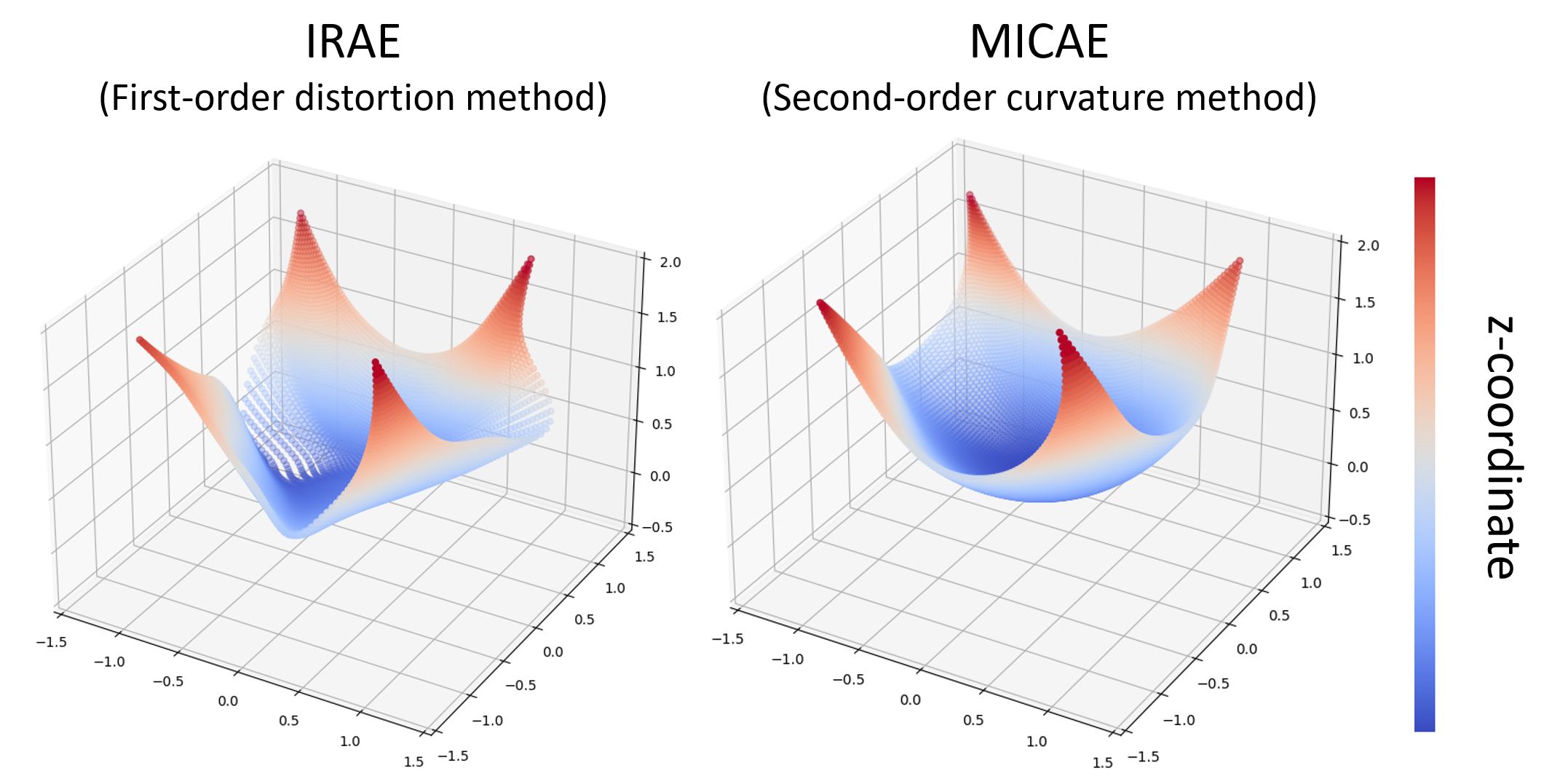}
    \vspace{-5pt}
    \caption{Manifold learning results of the first-order distortion minimization method, IRAE, and second-order (intrinsic) curvature minimization method, MICAE, for noise level 0.2.}
    \label{fig:first_second}
\end{figure}

Figure~\ref{fig:first_second} shows that MICAE learns a more accurate manifold than IRAE. 
A potential explanation for this is that IRAE has an extra requirement of obtaining undistorted coordinates in addition to minimizing intrinsic curvature, which can make the task of the generator network more challenging compared to that of the MICAE. Meanwhile, we found both methods converge to cylinder-like manifolds as the regularization coefficient or weight parameter $\alpha$ increases. When the level of noise is low ($\leq 0.1$), the two methods did not show a significant difference.

\subsection{Human Skeleton Data}
\label{subsec:skeleton}
In this section, we evaluate our minimum curvature autoencoders with the human skeleton pose data adopted from the NTU RGB+D dataset~\cite{shahroudy2016ntu}. A human pose skeleton data consists of 25 three-dimensional key points and thus is considered a 75-dimensional vector. There are 60 action classes (e.g., drinking water, brushing teeth), and each action data consists of a sequence of skeleton poses. We use randomly selected 800, 200, and 9000 skeleton poses for each action class as training, validation, and test data. We add Gaussian noises of standard deviations 0.05 and 0.1 to the training data. We use two-layer fully connected neural networks (512 nodes per layer) for both encoder and decoder with ELU activation functions, and the latent space dimension is 8. 

\begin{table}[!h]
\scriptsize
\caption{Averages and standard errors of the skeleton test data set reconstruction MSEs with Gaussian noise of standard deviations 0.05, 0.1, the lower, the better. The best results are marked in red. Those that show comparable performance to the best results are marked in bold (if the mean errors are included in the error bars of the best results). The numbers are written in units of $10^{-3}$.}
\label{tab:humanpose}
\begin{center}
\begin{tabular}{c|ccccc}
Noise & \multicolumn{2}{c}{0.05} & \multicolumn{2}{c}{0.1}
\\
Metric & \textit{clean2clean}  & \textit{corrupt2clean} & \textit{clean2clean} & \textit{corrupt2clean}
\\ \hline \hline
\textit{graph-free} \\ 
AE  & 7.52 $\pm$ 0.18 & 9.03 $\pm$ 0.16 & 15.4 $\pm$ 0.18 & 19.4 $\pm$ 0.18  
\\
VAE & 5.56 $\pm$ 0.18 & 6.74 $\pm$ 0.18 & 13.6 $\pm$ 0.24 & 19.1 $\pm$ 0.26 
\\
CAE & 2.32 $\pm$ 0.08 & 2.60 $\pm$ 0.08 & 2.48 $\pm$ 0.08 & 3.54 $\pm$ 0.08
\\
DAE & 3.09 $\pm$ 0.12 & 3.26 $\pm$ 0.12 & 5.55 $\pm$ 0.17 & 6.48 $\pm$ 0.18  
\\
RCAE & 3.77 $\pm$ 0.14  & 3.99 $\pm$ 0.14 & 6.51 $\pm$ 0.17 & 7.17 $\pm$ 0.19
\\
IRAE & \textcolor{red}{\textbf{2.18 $\pm$ 0.08}} & \textcolor{red}{\textbf{2.46 $\pm$ 0.08}} & \textbf{2.44 $\pm$ 0.09} & 3.55 $\pm$ 0.09 \\
MICAE & \textbf{2.20 $\pm$ 0.08} & \textbf{2.48 $\pm$ 0.08}  & \textcolor{red}{\textbf{2.37 $\pm$0.08}} & \textcolor{red}{\textbf{3.45 $\pm$ 0.08}}  
\\
MECAE & 2.29 $\pm$ 0.08 & 2.56 $\pm$ 0.08 & \textbf{2.40 $\pm$ 0.08} & \textbf{3.46 $\pm$ 0.08}
\\
\hline 
\textit{graph-based} \\ 
NRAE-L & \textbf{2.19 $\pm$ 0.08} & \textbf{2.46 $\pm$ 0.08} & 2.73 $\pm$ 0.08 & 3.79 $\pm$ 0.08 \\ 
NRAE-Q & 2.51 $\pm$ 0.10 & 2.92 $\pm$ 0.11 & 4.00 $\pm$ 0.11 & 5.93 $\pm$ 0.12  
\end{tabular}
\end{center}
\end{table}

\begin{figure*}[!t]
    \centering
    \includegraphics[width=0.8\linewidth]{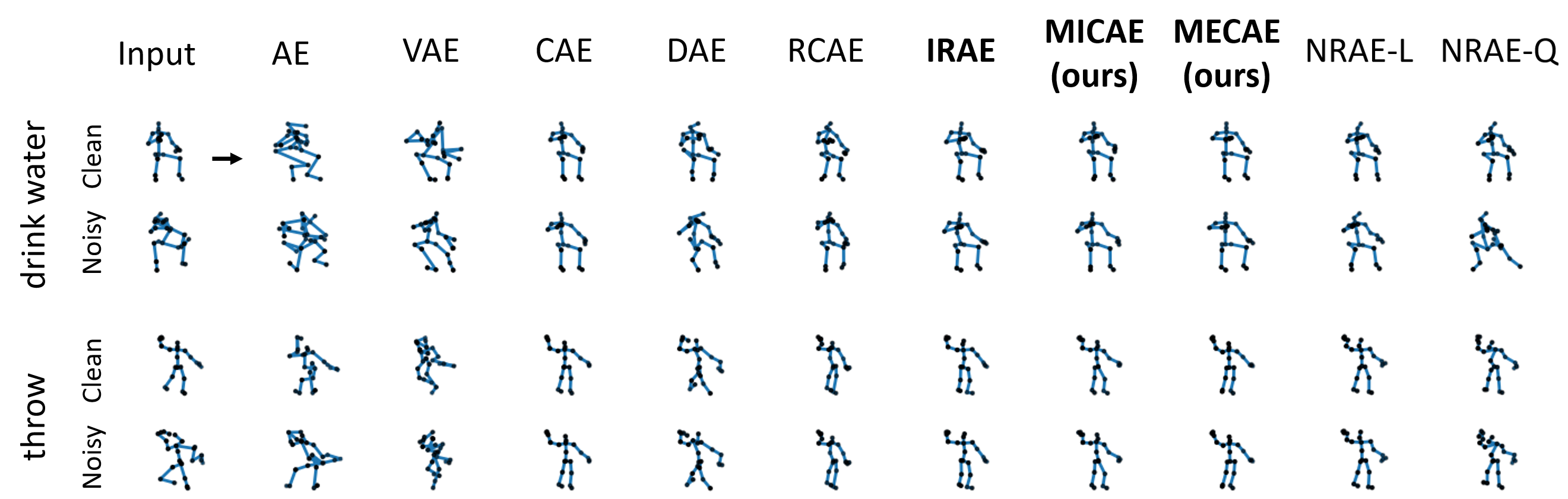}
    \caption{Clean and corrupted test data reconstruction results. Exemplar skeletons are selected from the ``drink water'' and ``throw'' action classes (noise level is 0.1). Methods that show qualitatively best reconstruction results are marked in bold (IRAE, MICAE, MECAE).}
    \label{fig:NTU_results1}
\end{figure*}

Table~\ref{tab:humanpose} shows the averages, and standard errors of the clean and corrupted test data set reconstruction MSEs over 60 different action classes, the lower, the better. 
IRAE produces the lowest reconstruction error when the noise level is 0.05, and MICAE produces the lowest reconstruction error when the noise level is 0.1. 
Overall, our minimum curvature autoencoders produce lower reconstruction errors, where MICAE slightly outperforms MECAE.
Figure~\ref{fig:NTU_results1} shows some example reconstruction results, where IRAE, MICAE, and MECAE seem to be most effective at mitigating noises and producing accurate skeletons, followed by CAE and NRAEs.

\begin{figure}[!t]
    \centering
    \includegraphics[width=1\linewidth]{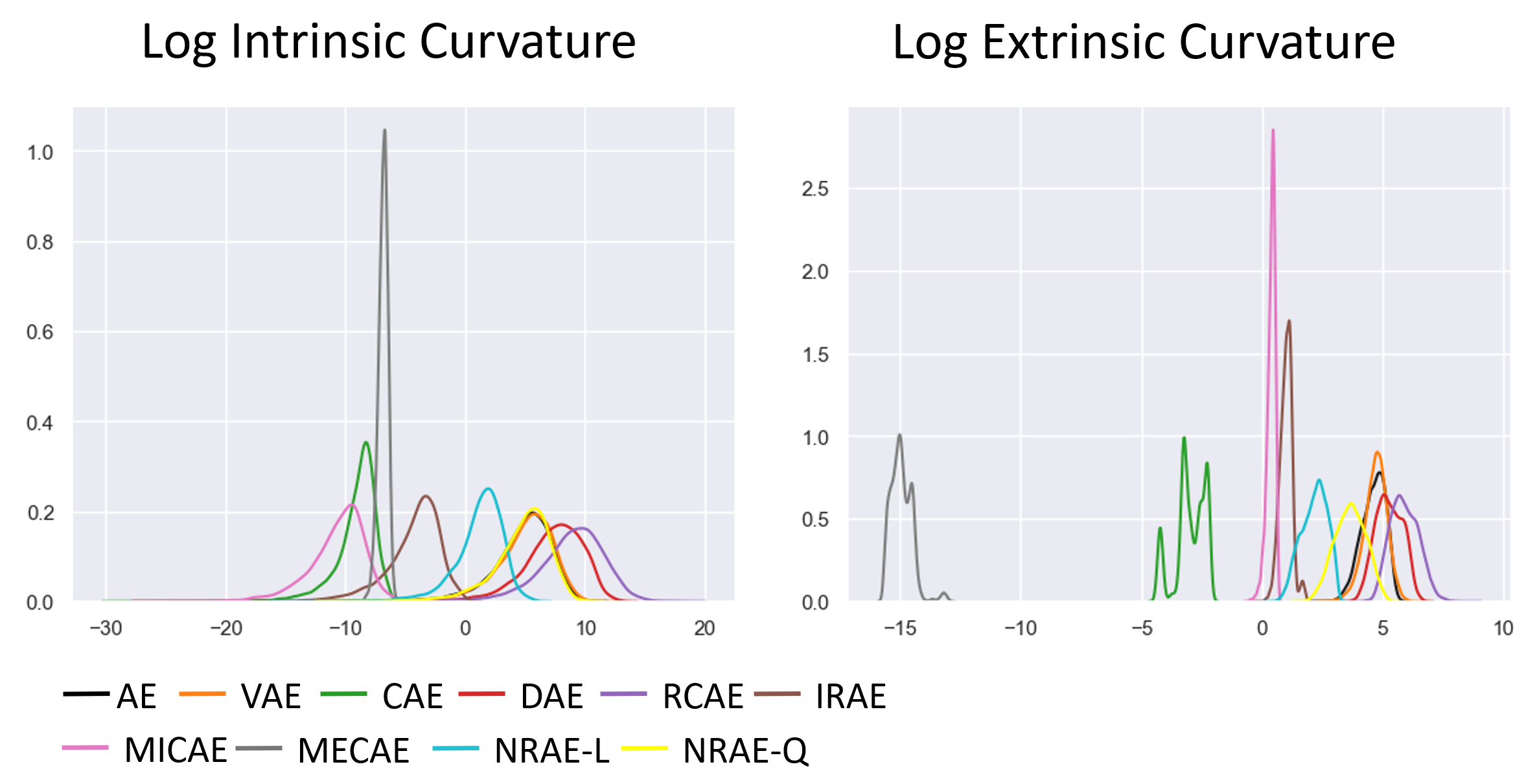}
    \vspace{-10pt}
    \caption{Density plots for the Log Intrinsic Curvature and Log Extrinsic Curvature, where the autoencoders are trained with data from the ``throw'' action class (noise level is 0.1).}
    \label{fig:NTU_results}
     \vspace{-5pt}
\end{figure}

Figure~\ref{fig:NTU_results} shows the Log Intrinsic Curvature and Log Extrinsic Curvature density plots of the learned manifolds. We train autoencoders with the corrupted training data from the ``throw" action class (noise level is 0.1) and compute the density plots with the test data set. 
As expected, MECAE produces the smallest extrinsic curvature and MICAE produces the smallest intrinsic curvature. 
One interesting observation is that other regularization methods, IRAE, and NRAE-L, which show some decent performance in mitigating the effects of noise, also reduce the curvature of the manifold, although their loss functions do not explicitly involve the curvature terms.  

\section{Conclusions}
We have proposed a class of curvature-based regularization terms for deep generative model learning.
Formulas for measuring both intrinsic and extrinsic curvature, which are independent of the choice of coordinates,  
have been developed for data manifolds of arbitrary dimensions embedded in higher-dimensional Euclidean space.
Since the formulas require a lot of computations as they require the evaluation of second-order derivatives,
we have developed efficient ways to approximate these curvatures.
Experiments that include noisy motion capture data have shown that curvature-based regularization methods are more effective for noise-robust manifold learning than the existing autoencoder methods, where using intrinsic curvature measures slightly outperforms using extrinsic curvature measures.

\textbf{Limitations and future research directions:}
While approximate formulas allow for the computation of intrinsic and extrinsic curvature terms, their calculations can still be quite time-consuming. Specifically, computing the inverse of the pullback Riemannian metric $G(z)^{-1}$ requires a significant amount of time and becomes even more challenging with increasing latent space dimensionality. Therefore, it is important to develop methods for even more approximating curvature computations that can drastically reduce the computation time, thus enabling their use in data and latent spaces of much larger dimensions.

While we have assumed the ambient data space is Euclidean, a growing number of problems involve non-Euclidean data (e.g., diffusion tensor data~\cite{fletcher2007riemannian}, point cloud data~\cite{lee2022statistical}).  
To develop minimum curvature deep generative models for non-Euclidean data, we need to generalize the intrinsic and extrinsic curvature measures for manifolds embedded in higher-dimensional Riemannian manifolds. 

\section*{Acknowledgements}
Yonghyeon Lee is the beneficiary of an individual grant from CAINS supported by a KIAS Individual Grant (AP092701) via the Center for AI and Natural Sciences at Korea Institute for Advanced Study.
Frank C. Park is supported in part by NRF-MSIT grant RS-2023-00208052, IITP-MSIT grant 2021-0-02068 (SNU AI Innovation Hub), KIAT-MOTIE grant P0020536 (HRD Program for Industrial Innovation), IITP-MSIT grant 2022-0-00480 (Training and Inference Methods for Goal-Oriented AI Agents), SNU-AIIS, SNU-IAMD, SNU BK21+ Program in Mechanical Engineering, SNU Institute for Engineering Research. 

\bibliography{example_paper}
\bibliographystyle{icml2023}

\newpage
\appendix
\onecolumn

\section*{Appendix}

\section{Coordinate-Invariance of the Extrinsic Curvature Measure}
\label{appendix:cioecm}
In this section, we show that (i) the representation of $T_{x} \mathcal{M}$, $E(J_f)=J_f(J_f^T J_f)^{-1}J_f^T$, is coordinate-invariant and (ii) the local extrinsic curvature measure, $\frac{1}{2}\text{Tr}(\sum_{i,j=1}^{m} (G^{-1})_{ij} \big( \frac{\partial }{\partial z^i} T \big)^T \frac{\partial }{\partial z^j} T)$, is coordinate-invariant. Given a decoder $f:z\in\mathbb{R}^{m} \mapsto x\in\mathcal{M}$, consider a coordinate transformation $h:z\in\mathbb{R}^{m} \mapsto z'\in \mathbb{R}^{m}$ and the transformed decoder $f':=f \circ h^{-1}:z'\in\mathbb{R}^{m} \mapsto x\in \mathcal{M}$. The Jacobian of $f$, $J_f$, is transformed as follows: 
\begin{equation}
    J_f \mapsto J_{f'} = \frac{\partial f'}{\partial z'} = \frac{\partial f}{\partial z} \frac{\partial h^{-1}}{\partial z'} = J_f H, 
\end{equation}
where $H=\frac{\partial h^{-1}}{\partial z'}$ is an $m \times m$ invertible matrix at $z$.  Therefore the representation $E(J_f)$ is coordinate-invariant:
\begin{equation}
    E(J_f) \mapsto E(J_{f'}) = J_{f'}(J_{f'}^T J_{f'})^{-1}J_{f'}^T = J_f H (H^T J_f^T J_f H)^{-1} H^T J_f^T = E(J_f). 
\end{equation}

Next, to show the coordinate-invariance of the extrinsic curvature, firstly, investigate how the Riemannian metric $G$ is transformed:
\begin{equation}
    G \mapsto G' = J_{f'}^T J_{f'} = H^T J_f^T J_f H = H^T G H.
\end{equation}
Since $T'=T$ and $\frac{\partial }{\partial z^{'i}}T = \sum_{a=1}^{m} H_{ai} \frac{\partial }{\partial z^a} T$, the curvature measure is coordinate-invariant:
\begin{align}
    \text{Tr}(\sum_{i,j=1}^{m} (G^{-1})_{ij} \big( \frac{\partial }{\partial z^i} T \big)^T \frac{\partial }{\partial z^j} T) & \mapsto \text{Tr}(\sum_{i,j=1}^{m} (G^{'-1})_{ij} \big( \frac{\partial }{\partial z^{'i}} T' \big)^T \frac{\partial }{\partial z^{'j}} T') \nonumber \\
    &= \text{Tr} \Big( \sum_{i,j=1}^{m} (H^T G H)^{-1}_{ij} \big( \sum_{a=1}^{m} H_{ai}\frac{\partial }{\partial z^a} T \big)^T \sum_{b=1}^{m} H_{bj} \frac{\partial }{\partial z^{j}} T \Big) \nonumber \\
    &= \text{Tr} \Big( \sum_{i,j,k,l=1}^{m} (H^{-1})_{ik} (G^{-1})_{kl}(H^{-T})_{lj} \big( \sum_{a=1}^{m} H_{ai} \frac{\partial }{\partial z^a} T \big)^T \sum_{b=1}^{m} H_{bj} \frac{\partial }{\partial z^{b}} T \Big) \nonumber\\
    &= \text{Tr} \Big( \sum_{i,j,k,l,a,b=1}^{m} (H^{-1})_{ik} (H^{-T})_{lj} H_{ai} H_{bj} (G^{-1})_{kl} \big(   \frac{\partial }{\partial z^a} T \big)^T \frac{\partial }{\partial z^{b}} T \Big) \nonumber \\ 
    &= \text{Tr} \Big( \sum_{k,l,a,b=1}^{m} \delta_{ak}\delta_{bl} (G^{-1})_{kl} \big(   \frac{\partial }{\partial z^a} T \big)^T \frac{\partial }{\partial z^{b}} T \Big) \nonumber \\
    &= \text{Tr} \Big( \sum_{a,b=1}^{m} (G^{-1})_{ab} \big(   \frac{\partial }{\partial z^a} T \big)^T \frac{\partial }{\partial z^{b}} T \Big),
\end{align}
where $\delta_{ij}$ is the Kronecker delta symbol.

\newpage
\section{Local Intrinsic Curvature Estimation}
\label{appendix:licef}
In this section, we provide simplified estimation formulas for the second, third, and fourth terms in (\ref{eq:est_ic_measure}), and summarize the strategy for computing (\ref{eq:ICmeasure}). 

\textbf{Second Term:} Plugging (\ref{eq:christoffel}) in the second term of (\ref{eq:est_ic_measure}), we get 
\begin{equation}
    \partial_j \Gamma^{a}_{ai}\tilde{v}^{i}v^j = \frac{1}{2} \partial_j \big( g^{a\lambda} (\tilde{v}^{i}\partial_i (g_{\lambda a})v^j + \partial_a (g_{\lambda i}\tilde{v}^{i})v^j - \partial_{\lambda} (g_{ai}\tilde{v}^{i})v^j ) \big).
\end{equation}
Since $\tilde{v}^ig_{ij} = v_i$, the second and third terms vanish, and then the second term of (\ref{eq:est_ic_measure}) is simplified to
\begin{equation}
    \partial_j \Gamma^{a}_{ai}\tilde{v}^{i}v^j = \frac{1}{2} v^j \partial_j \big( g^{a\lambda} (\tilde{v}^{i}\partial_i (g_{\lambda a})) = \frac{1}{2} v^j \partial_j \text{Tr} \big( G^{-1} \tilde{v}^{i}\partial_i G \big).
\end{equation}
Using Hutchinson’s trace estimator again, 
\begin{equation}
    \frac{1}{2} v^j \partial_j \text{Tr} \big( G^{-1} \tilde{v}^{i}\partial_i G \big) \approx \frac{1}{2}  (v \cdot \nabla)( w^T G^{-1} (\tilde{v} \cdot \nabla) (Gw)),
\end{equation}
for $w\in \mathcal{N}(0,I)$. 

\textbf{Third Term:} Plugging (\ref{eq:christoffel}) in the third term of (\ref{eq:est_ic_measure}), we get 
\begin{equation}
    \Gamma^{a}_{ab}\Gamma^{b}_{ij}\tilde{v}^{i}v^j = \frac{1}{2} g^{a\lambda} (\partial_b g_{a\lambda} + \partial_a g_{b\lambda} - \partial_{\lambda} g_{ab}) \times \frac{1}{2} g^{b\gamma} (v^j\partial_j (\tilde{v}^i g_{i\gamma}) + \tilde{v}^i \partial_i (v^j g_{j\gamma}) - \partial_{\gamma} (\tilde{v}^i g_{ij} v^j)).
\end{equation}
Since $\tilde{v}^ig_{ij} = v_i$, the first and third terms in the second bracket vanish, and then the third term of (\ref{eq:est_ic_measure}) is simplified to
\begin{equation}
    \Gamma^{a}_{ab}\Gamma^{b}_{ij}\tilde{v}^{i}v^j = \frac{1}{4} g^{a\lambda} (\partial_b g_{a\lambda} + \partial_a g_{b\lambda} - \partial_{\lambda} g_{ab}) \times g^{b\gamma} (\tilde{v}^i \partial_i (v^j g_{j\gamma})).
\end{equation}
To further simplify, we denote the term after the multiplication sign $g^{b\gamma} (\tilde{v}^i \partial_i (v^j g_{j\gamma}))$ by 
\begin{equation}
    V^b : = g^{b\gamma} (\tilde{v}^i \partial_i (v^j g_{j\gamma})) = \big( G^{-1} (\tilde{v}\cdot \nabla) (Gv)  \big)^b.
\end{equation}
Then the third term is simplified to 
\begin{align}
    \Gamma^{a}_{ab}\Gamma^{b}_{ij}\tilde{v}^{i}v^j &= \frac{1}{4} \big( g^{a\lambda} V^b\partial_b g_{a\lambda} + g^{a\lambda}\partial_a V^b g_{b\lambda} - g^{a\lambda} \partial_{\lambda} g_{ab}V^b \big) \nonumber \\
    &= \frac{1}{4} \big(\text{Tr}(G^{-1} (V \cdot \nabla)(G) ) \big),
\end{align}
since the second and third terms cancel each other out. Using Hutchinson’s trace estimator again, 
\begin{equation}
    \frac{1}{4} \big(\text{Tr}(G^{-1} (V \cdot \nabla)(G) ) \big) \approx \frac{1}{4} \big(w^T G^{-1} (V \cdot \nabla)(Gw) \big),
\end{equation}
for $w\in \mathcal{N}(0,I_m)$. 

\textbf{Fourth Term:} Plugging (\ref{eq:christoffel}) in the fourth term of (\ref{eq:est_ic_measure}), we get 
\begin{equation}
    \Gamma^{a}_{ib}\Gamma^{b}_{aj} \tilde{v}^i v^j = \frac{1}{2} \big( g^{a\lambda}( \tilde{v}^{i} \partial_i g_{\lambda b} + \partial_b (g_{\lambda i}\tilde{v}^{i}) - \partial_{\lambda} (g_{b i}\tilde{v}^{i}) ) \big) \times \frac{1}{2} \big( g^{b\gamma} ( v^j\partial_j g_{\gamma a} +\partial_a (g_{\gamma j}v^j) - \partial_{\gamma} (g_{aj}v^j) ) \big).
\end{equation}
Since $\tilde{v}^ig_{ij} = v_i$, the second and third terms in the first bracket vanish, and then the fourth term of (\ref{eq:est_ic_measure}) is simplified to
\begin{equation}
    \Gamma^{a}_{ib}\Gamma^{b}_{aj} \tilde{v}^i v^j = \frac{1}{4} \big( g^{a\lambda}\tilde{v}^{i} \partial_i g_{\lambda b}\big) \times \big( g^{b\gamma} ( v^j\partial_j g_{\gamma a} +\partial_a (g_{\gamma j}v^j) - \partial_{\gamma} (g_{aj}v^j) ) \big).
\end{equation}
To further simplify, we denote the term before the multiplication sign $g^{a\lambda} (\tilde{v}^i \partial_i (g_{\lambda b}))$ by 
\begin{equation}
    W^a_b : = g^{a\lambda} (\tilde{v}^i \partial_i (g_{\lambda b})) = \big( G^{-1} (\tilde{v} \cdot \nabla)(G)\big)^a{_b}.
\end{equation}
We note that $V=Wv$. Then the fourth term is simplified to 
\begin{align}
    \Gamma^{a}_{ib}\Gamma^{b}_{aj} \tilde{v}^i v^j &= \frac{1}{4} \big( W^a_b g^{b\gamma} ( v^j\partial_j g_{\gamma a} +\partial_a (g_{\gamma j}v^j) - \partial_{\gamma} (g_{aj}v^j) ) \big) \\ 
    &= \frac{1}{4} \big( 
        \text{Tr}(WG^{-1}(v\cdot\nabla)(G)) + \text{Tr}(WG^{-1}\nabla(Gv) -  \text{Tr}(G^{-1} W^T \nabla(Gv)) 
    \big).
\end{align}
Using Hutchinson’s trace estimator again, 
\begin{equation}
    \Gamma^{a}_{ib}\Gamma^{b}_{aj} \tilde{v}^i v^j = \frac{1}{4} \big( 
        w^T WG^{-1}(v\cdot\nabla)(Gw) + w^T (WG^{-1}-G^{-1} W^T)(w\cdot\nabla)(Gv)
    \big).
\end{equation}
for $w\in \mathcal{N}(0,I_m)$. 

Combining all four terms, finally we get the following local intrinsic curvature estimation formula:
\begin{align}
\label{eq:eis_formula_all}
    \text{EIC}(z;f) &= \Big( \frac{1}{2} (w \cdot \nabla) (w^T G^{-2} \textcolor{red}{(v\cdot \nabla) (Gv)}) \nonumber \\
    & - \frac{1}{2} (v \cdot \nabla) (w^T G^{-2} \textcolor{blue}{(v\cdot \nabla) (Gw)}) \nonumber \\
    & + \frac{1}{4} (w^T G^{-3} \textcolor{ao(english)}{(v\cdot \nabla)(G)} \textcolor{blue}{(v\cdot\nabla)(Gw)}) \nonumber \\
    & - \frac{1}{4} (w^T G^{-2} \textcolor{ao(english)}{(v\cdot \nabla)(G)} G^{-1} \textcolor{blue}{(v\cdot\nabla)(Gw)}) \nonumber \\
    & - \frac{1}{4} (w^T G^{-2} \textcolor{ao(english)}{(v\cdot \nabla)(G)} G^{-1} \textcolor{arylideyellow}{(w\cdot\nabla)(Gv)}) \nonumber \\ 
    & + \frac{1}{4}(w^T G^{-1} \textcolor{ao(english)}{(v\cdot \nabla)(G)} G^{-2} \textcolor{arylideyellow}{(w\cdot\nabla)(Gv)}) \Big)^2,
\end{align}
where $v,w \in \mathcal{N}(0, I)$ and $G=J_f^T J_f$; it can be computed by using the Jacobian-vector and vector-Jacobian products multiple times.

\newpage
\section{Algorithmic Details}
\label{appendix:algo_details}

In this section, we describe algorithms for MICAE and MECAE.  
Given a decoder $f:\mathbb{R}^{m} \to \mathbb{R}^{D}$, at a latent point $z\in\mathbb{R}^{m}$, denote the estimated intrinsic curvature by $\textrm{EIC}(z; f)$ (\ref{eq:eis_formula_all}) and the estimated extrinsic curvature by $\textrm{EEC}(z;f)$ (\ref{eq:esti_extrinsic_curv}).  
Let $g_{\phi}:\mathbb{R}^{D} \to \mathbb{R}^{m}$ be a parameterized encoder and $f_{\theta}:\mathbb{R}^{m}\to\mathbb{R}^{D}$ be a parameterized decoder, where $\theta,\phi$ are learnable parameters.

The training of MICAE and MECAE is as simple as the vanilla autoencoder, where the only difference is the added regularization term in the loss function. The new loss functions have the following form: 
\begin{equation}
    \mathbb{E}_{x \sim p_{\textrm{data}}} [\|f_{\theta} (g_{\phi}(x)) - x\|^2] + \alpha \ \mathbb{E}_{x \sim p_{\textrm{data}}} [ \textrm{EC}(g_{\phi}(x); f_{\theta})],
\end{equation}
where $\alpha$ is the regularization coefficient, $p_{\textrm{data}}$ is the empirical training data distribution, and $\textrm{EC}(z;f)$ is either $\textrm{EIC}(z;f)$ or $\textrm{EEC}(z;f)$. 
In practice, $\mathbb{E}_{x \sim p_{\textrm{data}}}$ is replaced by $\frac{1}{N}\sum_{i=1}^{N}$ with the set of training data $\{x_i\}_{i=1}^{N}$.

\section{Experimental Details}
\label{appendix:exp_details}

\textbf{Latent Space Augmentation:} Sometimes, a simple latent space data augmentation technique can improve the manifold learning performance. Following~\cite{lee2022regularized, chen2020learning}, we use the modified mix-up data-augmentation method with a parameter $\eta >0$. The encoded data is augmented by $z = \delta z_1 +  (1-\delta)z_2$ where $z_i, i = 1, 2$ are randomly sampled two encoded data and $\delta$ is uniformly sampled from $(-\eta, 1 + \eta)$ (we set $\eta= 0.2$). Note that we use this technique for synthetic data. 

\subsection{Synthetic Data}
For both two-dimensional and one-dimensional manifold examples, we use three-layer fully
connected neural networks (512 nodes per layer) for both encoder and decoder with ELU activation functions. We use Adam optimizer with a learning rate of 0.00001, and the number of training epochs is 10000.
The regularization coefficient for the minimum curvature autoencoders is searched in (0.0001, 0.001, 0.01, 0.1, 1, 10), and the one that produces the smallest (clean2clean) test reconstruction error is reported. 

For the two-dimensional manifold example, when computing the intrinsic and extrinsic curvatures for the density plots, we did not use the curvature estimation formulas, rather compute them precisely using the original formulas, which is feasible since both the data and latent space dimension are low. 
Table~\ref{tab:pert_syn} shows the runtimes per epoch with the two-dimensional manifold example for a batch size of 100.
\begin{table}[!h]
\scriptsize
\vspace{-10pt}
\caption{The runtimes per epoch for a batch size of 100 (where we use the GeForce RTX 3090).}
\label{tab:pert_syn}
\vspace{5pt}
\begin{center}
\begin{tabular}{cccc}
\hline
AE & MICAE & MECAE  \\ 
\hline
0.003 s & 0.087 s & 0.029 s \\
\hline
\end{tabular}
\end{center}
\vspace{-10pt}
\end{table}

\subsection{Motion Capture Data}
We use two-layer fully connected neural networks (512 nodes per layer) for both encoder and decoder with ELU activation functions. We use Adam optimizer with a learning rate of 0.001, and the number of training epochs is 5000.
We search the hyperparameters for each method over a wide enough range and report the results of the one that produces the smallest (clean2clean) validation reconstruction error. 

When computing the intrinsic and extrinsic curvatures for the density plots, we did not use the curvature estimation formulas, rather compute them precisely using the original formulas, which is feasible since both the data and latent space dimension are low. 
Table~\ref{tab:pert_ntu} shows the runtimes per epoch for a batch size of 100.
\begin{table}[!h]
\scriptsize
\vspace{-10pt}
\caption{The runtimes per epoch for a batch size of 100 (where we use the GeForce RTX 3090).}
\label{tab:pert_ntu}
\vspace{5pt}
\begin{center}
\begin{tabular}{cccccccccc}
\hline
AE & VAE & CAE & DAE & RCAE & IRAE & MICAE & MECAE & NRAE-L & NRAE-Q \\ 
\hline
0.0130 s & 0.0198 s & 0.0267 s & 0.0133 s & 0.0299 s & 0.0353 s & 0.4600 s & 0.1523 s & 0.0280 s & 0.0459 s \\
\hline
\end{tabular}
\end{center}
\vspace{-10pt}
\end{table}

\end{document}